\documentclass[11pt]{article}

\usepackage[preprint]{acl}

\usepackage{times}
\usepackage{latexsym}
\usepackage{listings}
\usepackage{xcolor}
\usepackage{enumitem}
\usepackage{graphicx}
\usepackage{tabularx}
\usepackage{fancyhdr} 
\usepackage{caption}
\usepackage{array}      
\usepackage{longtable}
\usepackage{amsmath}
\usepackage{amssymb}
\usepackage{booktabs}
\usepackage{multirow}
\usepackage[section]{placeins}
\usepackage{afterpage}


\lstdefinestyle{promptstyle}{
  basicstyle=\ttfamily\scriptsize,
  breaklines=true,
  breakatwhitespace=false,
  breakindent=0pt,
  postbreak=\mbox{$\hookrightarrow$\space},
  columns=fullflexible,
  keepspaces=true,
  showstringspaces=false,
  frame=single,
  framerule=0.4pt,
  rulecolor=\color{black!50},
  backgroundcolor=\color{black!3},
  xleftmargin=4pt,
  xrightmargin=4pt,
  aboveskip=4pt,
  belowskip=4pt,
  captionpos=b,
}
\usepackage[T1]{fontenc}

\usepackage[utf8]{inputenc}

\usepackage{microtype}

\usepackage{inconsolata}

\usepackage{graphicx}
\graphicspath{{./}{./image/}{./latex/image/}{./latex/}}

\title{EditCaption: Human-Refined SFT and HAE-DPO 
for Image Editing Instruction Synthesis
}

\author{
  \textbf{Xiangyuan Wang\textsuperscript{1,5*}}\quad
  \textbf{Honghao Cai\textsuperscript{2,5*}}\quad
  \textbf{Yunhao Bai\textsuperscript{1,5}}\quad
  \textbf{Chao Hui\textsuperscript{5}}\quad
  \textbf{Tianze Zhou\textsuperscript{3,5}}\\
  \textbf{Haohua Chen\textsuperscript{4,5}}\quad
  \textbf{Hao Shi\textsuperscript{3}}\quad
  \textbf{Yuling Wu\textsuperscript{2}}\quad
  \textbf{Yao Hu\textsuperscript{5}}\quad
  \textbf{Xu Tang\textsuperscript{5}}\quad\\
  \textbf{Yibo Chen\textsuperscript{5}}\quad
  \textbf{Wei Zhu\textsuperscript{5\dag}}\\[4pt]
  \textsuperscript{1}Peking University\quad
  \textsuperscript{2}The Chinese University of Hong Kong, Shenzhen\quad
  \textsuperscript{3}Tsinghua University\\
  \textsuperscript{4}Beihang University\quad
  \textsuperscript{5}Xiaohongshu Inc.\\[2pt]
  \textsuperscript{*}Equal contribution.\quad
  \textsuperscript{\dag}Corresponding author.
}

\begin{document}
\makeatletter
\setlength{\@fptop}{0pt}
\setlength{\@fpsep}{12pt}
\setlength{\@fpbot}{0pt plus 1fil}
\setlength{\@dblfptop}{0pt}
\setlength{\@dblfpsep}{12pt}
\setlength{\@dblfpbot}{0pt plus 1fil}
\makeatother
\maketitle
\begin{abstract}
High-quality source–target image pairs with precise editing instructions are essential for instruction-guided image editing, yet constructing such training triplets at scale remains costly. Recent pipelines often rely on vision-language models to synthesize editing instructions automatically, but we find that strong VLMs still struggle to describe visual transformations between image pairs. In particular, they exhibit three recurring failure modes: orientation inconsistency, viewpoint ambiguity, and missing fine-grained attributes. In a human evaluation on 400 image pairs, Several open-source VLM baselines produce critical-error rates above 47\%, making many synthesized instructions unsuitable for downstream training.

To address this, we propose \textbf{EditCaption}, a two-stage post-training pipeline for image editing instruction synthesis. First, We construct a 100K supervised fine-tuning dataset through GLM-based auto-captioning, EditScore filtering, and human refinement. Second, We collect 10K human-annotated preference pairs, where each rejected instruction is labeled with its primary error type and severity. Based on this dataset, we propose Hardness-Adaptive Error-Aware DPO (\textbf{HAE-DPO}), a task-adapted DPO objective that introduces an adaptive margin based on human-labeled severity, failure-mode type, and reference-model hardness.

Experiments across three benchmarks demonstrate that our 235B model with SFT+HAE-DPO achieves state-of-the-art performance among open-source and closed-models, scoring 4.720 on Eval-400, 4.672 on HQ-Edit, and 4.651 on ByteMorph-Bench—surpassing Gemini-3-Pro on all three. Human evaluation confirms critical error rates drop from 47.75\% to 17.50\%, with correct rates improving from 41.75\% to 70.25\%, surpassing Gemini-3-Pro(66.00\%).
\end{abstract}

\section{Introduction}

Instruction-guided image editing has emerged as a central paradigm in
controllable visual generation~\cite{brooks2023instructpix2pix,fu2023guiding}.
The quality of training triplets—source/target image pairs with precise
editing instructions—is critical to system performance. Constructing such
triplets at scale via manual annotation is prohibitively expensive, motivating
VLM-based automated synthesis~\cite{liu2023llava,bai2023qwenvl}. However,
this delegation rests on a flawed assumption: VLMs capable of rich
single-image captioning do not equally describe transformations between two
images. We identify three dominant failure modes: orientation inconsistency
(e.g., left/right reversal), viewpoint ambiguity, and missing fine-grained
detail. Human evaluation on 400 image pairs confirms that over 47\% of
baseline VLM outputs contain critical errors rendering them unusable for
downstream training—a persistent failure that scaling or prompt engineering
alone cannot resolve.

To address this, we propose \textbf{EditCaption}, a two-stage post-training
pipeline. First, we construct a 100K SFT dataset via GLM-based annotation,
EditScore~\cite{luo2025editscore} filtering, and human refinement. Second, we build 10K failure-labeled
preference pairs and introduce Hardness-Adaptive Error-Aware DPO
(\textbf{HAE-DPO}), a novel alignment objective that modulates the preference margin
jointly by error severity, failure-mode type, and model-perceived hardness—
going beyond the uniform-margin assumption of standard DPO~\cite{rafailov2023dpo}.
Experiments on three benchmarks show
our 235B model achieves 4.720 on Eval-400, surpassing Gemini-3-Pro (4.706),
with human evaluation confirming critical errors dropping from 47.75\% to
17.50\% and correct rates improving from 41.75\% to 70.25\%.

In summary, this paper makes three contributions:
\begin{itemize}
\item \textbf{Failure analysis.} We identify and quantify three dominant failure modes in VLM-based editing instruction synthesis: orientation inconsistency, viewpoint ambiguity, and missing fine-grained detail.

\item \textbf{Human-refined training data.} We construct a 100K SFT dataset and a 10K preference dataset, where preference pairs are explicitly annotated with error type and severity.

\item \textbf{HAE-DPO.} An adaptive-margin DPO variant that assigns stronger optimization pressure to severe, spatially important, and model-hard errors.
\end{itemize}

\section{Related Work}

\subsection{Instruction-Guided Image Editing}
Text-driven image editing has advanced rapidly with large-scale
diffusion models~\cite{rombach2022ldm}. Early methods such as
Prompt-to-Prompt~\cite{hertz2022prompt} and
Imagic~\cite{kawar2023imagic} achieve word-level edits through
cross-attention manipulation or text embedding optimization.
InstructPix2Pix~\cite{brooks2023instructpix2pix} instead learns
to follow free-form instructions from synthetic triplets generated
by GPT-3 and Stable Diffusion, while
MagicBrush~\cite{zhang2023magicbrush} provides a manually annotated
counterpart. SmartEdit~\cite{huang2024smartedit} further leverages MLLMs
to handle complex, compositional editing instructions. Ji et al.~\cite{ji2025instruction}
introduce a planning-and-reasoning framework that decomposes instructions into explicit
sub-goals before generation. FireEdit~\cite{zhou2025fireedit} targets fine-grained
spatial localization via region-aware VLM supervision. Recent benchmarks including
HQ-Edit~\cite{hui2024hq} and ByteMorph-Bench~\cite{chang2025bytemorph}
further emphasize instruction quality for spatially complex
transformations. Despite these advances, all such systems remain constrained by training
triplet quality. We address the preceding data-construction problem:
generating accurate natural-language instructions from source--target
image pairs.

\subsection{Vision-Language Models for Data Synthesis}
VLMs such as BLIP-2~\cite{li2023blip2},
InstructBLIP~\cite{dai2023instructblip},
LLaVA~\cite{liu2023llava}, and
Qwen-VL~\cite{bai2023qwenvl} can produce detailed, context-sensitive
image descriptions. However, these models are optimized for semantic
captioning of \emph{individual} images. When tasked with
characterizing transformations between image pairs, they exhibit
systematic failures in orientation, viewpoint, and fine-grained
detail---which we empirically quantify and address through
targeted post-training. We show that strong single-image captioners do not necessarily transfer to pairwise transformation captioning, where the model must compare two images and verbalize only the visual delta.

\subsection{Automated Data Quality and Filtering}
Automated triplet construction, exemplified by
InstructPix2Pix~\cite{brooks2023instructpix2pix}, is susceptible
to semantic misalignment and spatial inconsistencies, while manual
annotation~\cite{zhang2023magicbrush} is costly and limited in scale.
Automatic quality filtering has been explored through CLIPScore for
image-text alignment~\cite{hessel2021clipscore} and reward models in
RLHF pipelines~\cite{ouyang2022instructgpt}, but dedicated metrics for editing instruction evaluation are emerging but remain insufficient as a standalone solution for high-precision data construction. We leverage
EditScore~\cite{luo2025editscore}, an open-source reward model for
instruction-guided editing, as a data filtering mechanism.

\subsection{Preference Optimization}
RLHF~\cite{ouyang2022instructgpt} is widely used for aligning
language models with human intent. Direct Preference
Optimization~\cite{rafailov2023dpo} reformulates this as a
classification problem on preference pairs, eliminating the need
for an explicit reward model. While DPO has been extended to
multimodal settings, standard DPO treats all preference
pairs with a uniform margin, ignoring why a rejected
response fails. In our setting, human annotators explicitly label
each rejection with a failure mode (orientation, viewpoint, or
detail) and an error severity score, providing structured
supervision beyond binary preference. We leverage this supervision
in \textbf{HAE-DPO}, which modulates the preference margin
according to error severity, failure type, and model-perceived
learning difficulty—enabling more targeted alignment than
uniform-margin DPO.

\begin{figure*}[t]
  \centering
  \includegraphics[width=0.6\linewidth]{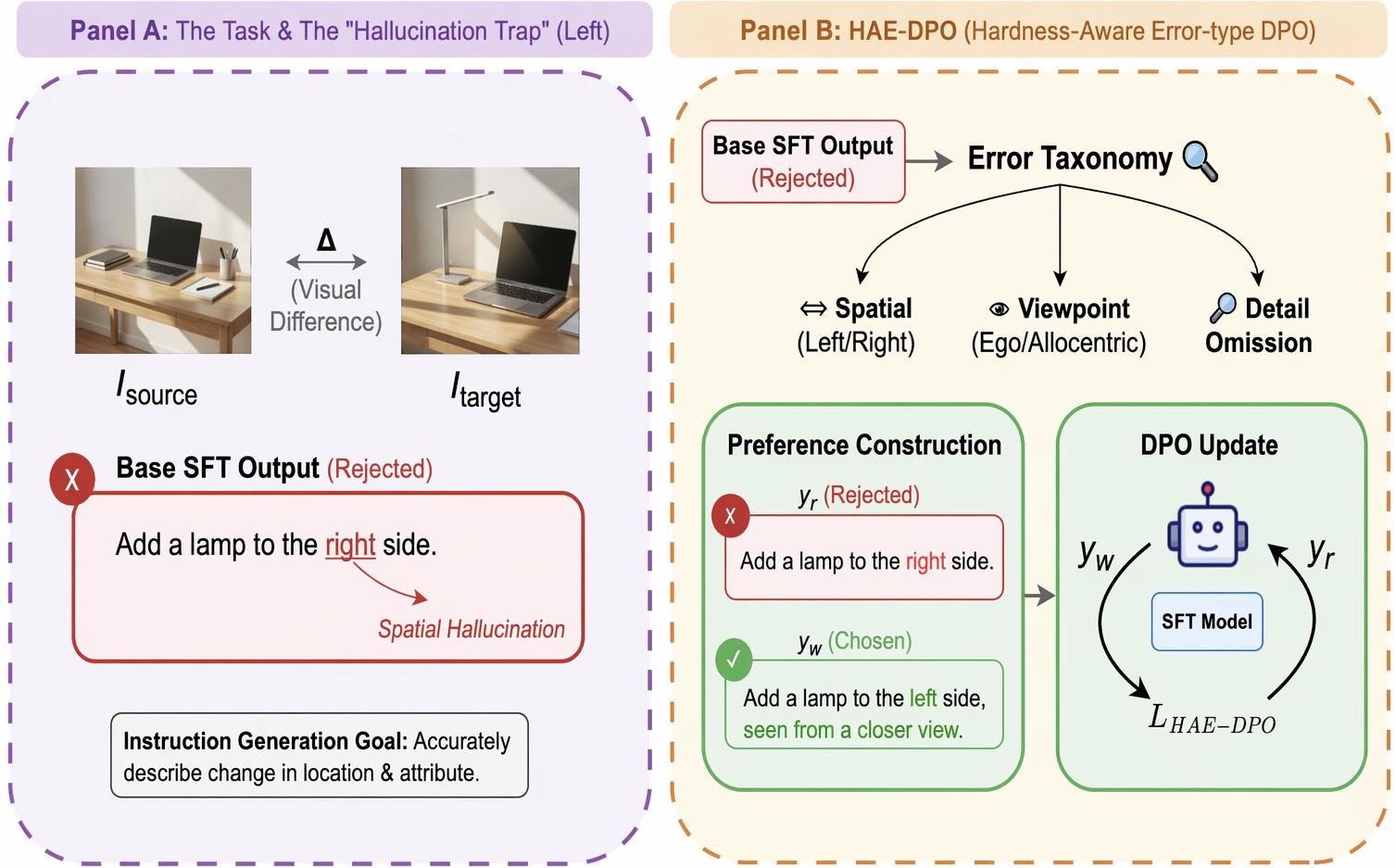}
  \caption{
    \textbf{Overview of our approach.}
    \textbf{(a) Task \& Failure Modes:} Given a source-target image pair,
  VLMs must generate accurate editing instructions but suffer from
  three systematic failures: directional confusion, viewpoint ambiguity,
  and missing fine-grained detail.
    \textbf{(b) HAE-DPO:} SFT model outputs serve as rejected samples;
human-corrected instructions as chosen samples. Each rejection is
annotated with an error type and severity score, enabling
HAE-DPO to modulate the preference margin beyond uniform DPO.
  }
  \label{fig:overview}
\end{figure*}
\section{Problem Description and Methodology}

\subsection{Problem Description}
We address the task of editing instruction synthesis: given a source
image $I_{\text{src}}$ and a target image $I_{\text{tgt}}$ representing a visual
transformation, the goal is to generate a natural language instruction
$y$ that precisely describes the editing operation required to transform
$I_{\text{src}}$ into $I_{\text{tgt}}$. Formally, we seek a model $f_{\theta}$ such that: $y = f_{\theta}(I_{\text{src}}, I_{\text{tgt}})$. 

where $y$ should satisfy three properties: (1) \textbf{Accuracy}: correctly describe all primary visual changes without hallucination; (2) \textbf{Completeness}: cover all primary changes without omission; (3) \textbf{Clarity}: be unambiguous and executable by a user without access to $I_{\text{tgt}}$.

This task is fundamentally harder than single-image captioning: the model must reason about the \textit{difference} between two images, requiring sensitivity to directional relationships, attribute-level changes, and viewpoint shifts. These failure modes are not symptoms of insufficient scale or prompt engineering—VLMs optimize for single-image token likelihood, a fundamentally different objective from spatial delta reasoning, and this misalignment persists across all tested models.

\subsection{Method Overview}
We propose \textbf{EditCaption}, a VLM fine-tuned through 
a two-stage post-training pipeline, as illustrated in 
Figure~\ref{fig:overview}. The two stages are designed to address complementary weaknesses: SFT establishes
the model's basic capacity to produce structured, accurate
instructions at scale, while DPO targets residual systematic errors
that persist after SFT.

\paragraph{Stage 1: SFT.} We train $f_{\theta}$ on a
curated dataset $\mathcal{D}_{\text{sft}} = \{(I_{\text{src}}^i, I_{\text{tgt}}^i, y^i)\}_{i=1}^{N}$ of
$N = 100K$ image pairs with human-refined instructions. The training
objective is standard cross-entropy:
\begin{equation}
\resizebox{\columnwidth}{!}{$
\mathcal{L}_{\text{SFT}} = -\mathbb{E}_{(I_{\text{src}}, I_{\text{tgt}}, y) \sim \mathcal{D}_{\text{sft}}} \left[ \sum_{t} \log p_{\theta}(y_t | y_{<t}, I_{\text{src}}, I_{\text{tgt}}) \right]
$}
\end{equation}

This stage instills the core generation capability: correctly
identifying and describing primary changes, resolving directional
relationships, and producing fine-grained attribute descriptions.

\paragraph{Stage 2: HAE-DPO.}
Standard DPO uses a fixed preference objective and does not explicitly distinguish whether a rejected instruction fails due to a critical direction reversal, a viewpoint ambiguity, or a minor detail omission.

We propose Hardness-Adaptive Error-Aware DPO (\textbf{HAE-DPO}),
which introduces a per-sample adaptive margin $m_i$:
\begin{equation}
\resizebox{\columnwidth}{!}{$
L_{\text{HAE-DPO}} = -\mathbb{E}\!\left[\log\sigma\!\left(
  \beta\log\frac{p_\theta(y_w|x)}{p_{\text{ref}}(y_w|x)}
  -\beta\log\frac{p_\theta(y_r|x)}{p_{\text{ref}}(y_r|x)}
  - m_i\right)\right]
$}
\end{equation}
where the adaptive margin combines three signals:
\begin{equation}
\resizebox{\columnwidth}{!}{$
m_i = \gamma_{st} s_ic(t_i) + \gamma_h\,\sigma\!\!\left(
        \frac{\log p_{\text{ref}}(y_r|x)
              -\log p_{\text{ref}}(y_w|x)}{\tau}
      \right)\\
    $}
\end{equation}
where $s_i \in [0, 0.5, 1]$ is the human-annotated error severity(P0/P1/P2), $c(t_i)$ is an
error-type weight that up-weights spatially critical failure modes (orientation inconsistency $= 0.2$, viewpoint ambiguity $= 0.15$, detail omission $= 0.1$), and $\gamma_{st}$, $\gamma_h$ are
balancing coefficients. The first term prioritizes human-perceived error importance: severe errors receive larger margins, and spatially critical error types such as orientation inconsistency are further up-weighted. The second term captures model-perceived hardness: if the reference SFT model assigns a higher normalized likelihood to the rejected instruction than to the corrected one, the pair is considered harder and receives a larger margin.

We apply this pipeline to two model scales: Qwen3-VL-32B and
Qwen3-VL-235B-A22B, using the same data and training procedure.
\section{Dataset Construction}
In this section, we describe the construction of both the supervised fine-tuning (SFT) dataset and the preference dataset for alignment. Our data pipeline is designed to progressively improve instruction quality, moving from large-scale automatic generation to human-refined supervision and finally to error-driven preference optimization.
\begin{figure*}[t]
  \centering
  \includegraphics[width=\linewidth]{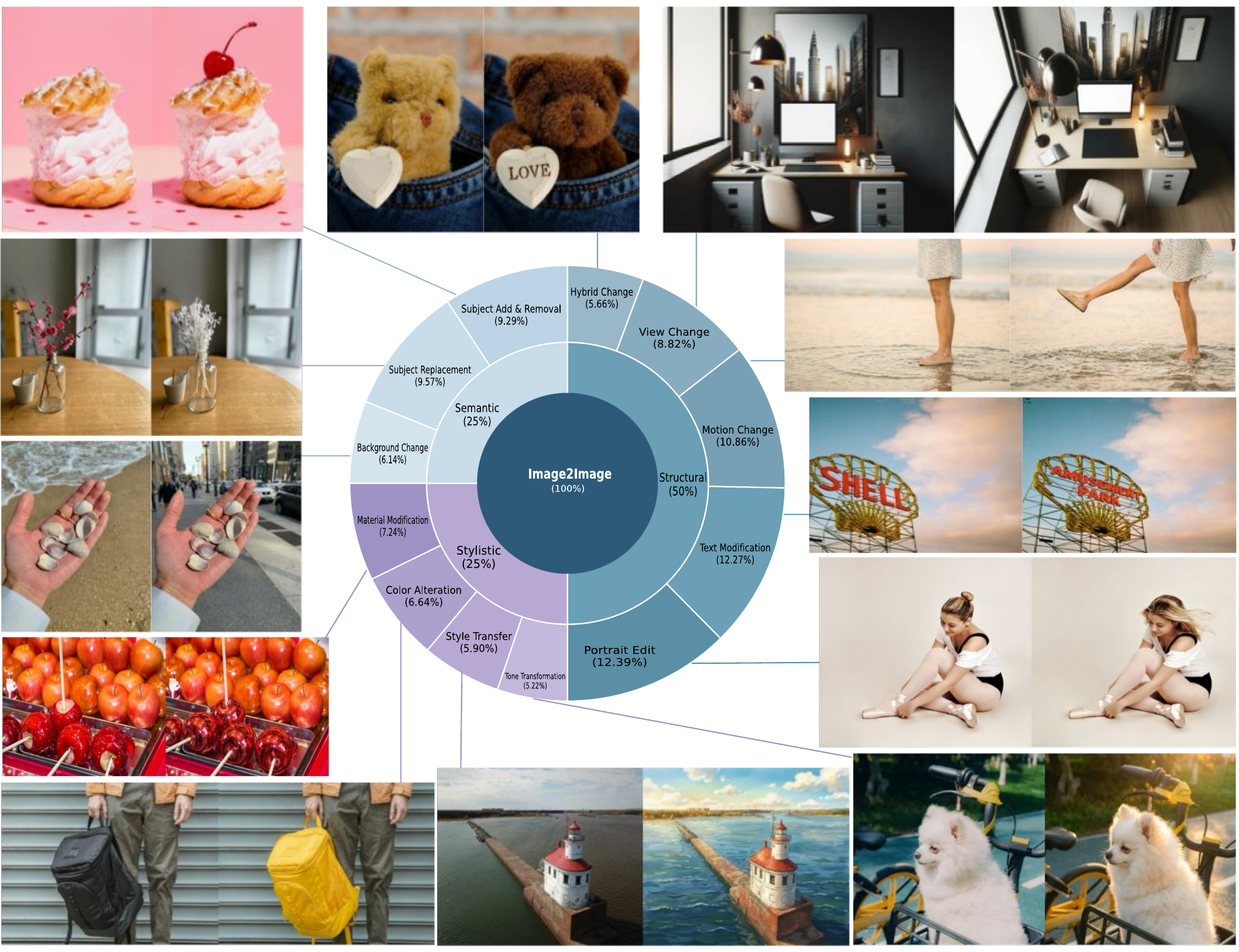}
  \caption{
    \textbf{Overview of Data Distribution.} Our training dataset is divided into three categories:
Semantic Editing (content-based modifications), Stylistic Editing (aesthetic adjustments), and Structural Editing(spatial arrangement and composition). Our data collection strategy ensures a balance of diversity and quality throughout the training process, providing comprehensive coverage and precise annotations to foster robust model training.
  }
  \label{fig:data}
\end{figure*}
\subsection{SFT Data Construction}
We begin with a large-scale collection of 150K image pairs sourced from our internal image editing pipeline, where each pair consists of a source image $I_\text{src}$ and a target image $I_\text{tgt}$ representing a visual transformation. To ensure broad coverage of real-world editing scenarios, the dataset is organized into three high-level categories based on the nature of the transformation:
\begin{itemize}
    \item \textbf{Semantic Editing} (25\%): Modifications to the \emph{content} of an image, including adding, removing, or replacing objects, and altering the background.

    \item \textbf{Stylistic Editing} (25\%): Changes to the \emph{visual style and aesthetic} of an image without altering its core content, such as color alteration, style transfer, tone transformation, and material modification. 

    \item \textbf{Structural Editing} (50\%): Changes to the \emph{spatial arrangement and composition} of the scene. This category includes control-intensive scenarios such as view change, motion change, and portrait change, as well as more complex cases such as text modification and hybrid transformations that require coordinated spatial and visual reasoning.
\end{itemize}
As illustrated in Figure~\ref{fig:data}, the 
dataset spans a wide range of real-world editing scenarios, 
with Structural Editing comprising the largest portion (50\%), 
reflecting its central role in the spatial reasoning challenges 
we target.
This taxonomy-guided data collection ensures that the SFT dataset provides sufficient supervision signal across the full spectrum of editing types, with particular emphasis on structural transformations that are most prone to systematic errors, particularly for structural and viewpoint changes.

\textbf{Initial Caption Generation}. To obtain initial supervision signals, we leverage GLM-4.5V to generate editing instruction captions for each image pair. This step enables scalable annotation but inevitably introduces noise due to imperfect visual understanding.

\textbf{Quality Scoring with EditScore}. We apply EditScore~\cite{luo2025editscore}, an open-source reward model that measures consistency between the generated instruction and the visual difference between $I_{\text{src}}$ and $I_{\text{tgt}}$, evaluating both editing success and degree of overediting. Samples with a composite score below 4.0 are discarded, retaining approximately 100K high-quality candidates.

\begin{figure}[t]
  \centering
  \includegraphics[width=\linewidth]{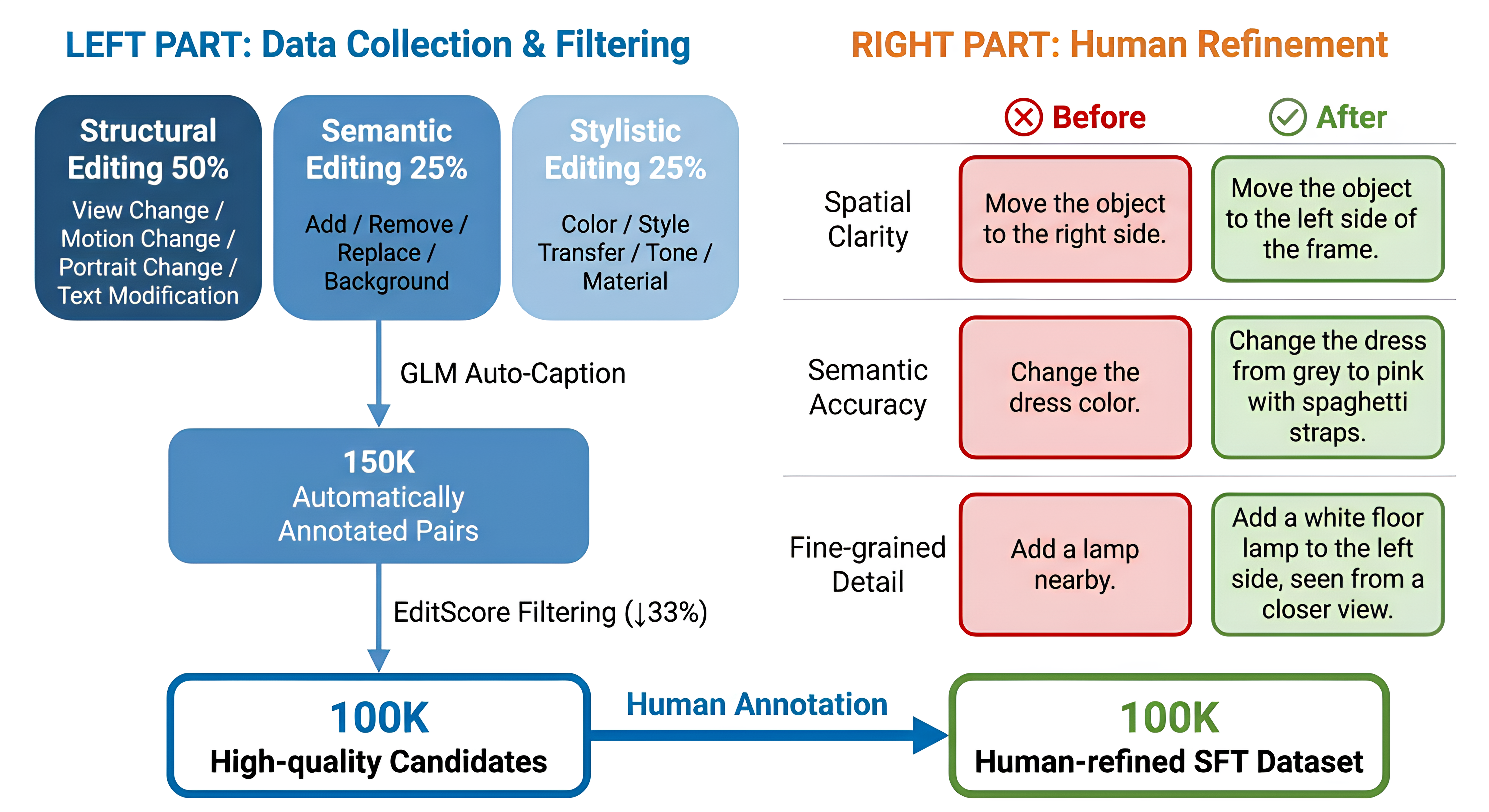}
  \caption{
    \textbf{The three-step SFT data construction pipeline.} 
  }
  \label{fig:sft}
\end{figure}

\textbf{Human Refinement.} From the filtered dataset, we select 100K 
samples for manual refinement. Annotators are instructed to ensure 
semantic accuracy (the instruction precisely reflects the visual change), 
resolve spatial relationships explicitly (e.g., left/right, relative 
positions), and enrich fine-grained attribute descriptions such as color, 
texture, and object properties. Figure~\ref{fig:sft} summarizes the 
complete three-step construction process.

\subsection{Preference Data Construction}
Despite strong performance after SFT, we observe that the model still exhibits systematic errors in instruction generation. Through empirical analysis, we identify three dominant failure modes:
\begin{itemize}
\item {\texttt{Orientation inconsistency}}: incorrect left/right or directional descriptions.
\item {\texttt{Viewpoint ambiguity}}: failure to correctly describe perspective or view transformations.
\item {\texttt{Lack of fine-grained detail}}: missing or vague attribute-level descriptions.
\end{itemize}
To address these failures, we generate instructions for 30K image pairs using the SFT-trained model, then select 10K samples exhibiting the three target failure modes as rejected examples. Human annotators provide corrected instructions as chosen samples, focusing on spatial consistency, viewpoint clarity, and fine-grained accuracy. Each rejected sample is further labeled with its primary failure mode (orientation inconsistency, viewpoint ambiguity, or lack of fine-grained detail) and severity (P0/P1/P2), which directly feed the per-sample margin $m_i$ in HAE-DPO. The resulting 10K preference pairs $(I_{\text{src}},I_{\text{tgt}},y_{\text{chosen}},y_{\text{rejected}})$ are used to train HAE-DPO for targeted alignment.
To prevent overfitting to majority error types, we balance the 10K preference
dataset across the three failure modes: orientation inconsistency (34.2\%),
viewpoint ambiguity (31.5\%), and lack of fine-grained detail (34.3\%).

\section{Experiments}
\subsection{Experimental Setup}
\subsubsection{Datasets}
We evaluate our method on three datasets, including one in-house test set and two public benchmarks. To provide a clear overview of dataset scale and characteristics, we summarize all evaluation datasets in Table 1.
\begin{table}[h]
  \centering
  \caption{Summary of evaluation datasets}
  \label{tab:freq}
  \resizebox{\columnwidth}{!}{%
  \begin{tabular}{p{2cm} p{1.5cm} p{6cm}}
    \toprule
    Dataset & \#Samples & Characteristics\\
    \midrule
    In-house Test Set & 400 & Diverse editing operations, covering spatial, viewpoint, and attribute changes\\
    \midrule
    ByteMorph-Bench & 600 & Non-rigid motion transformations with spatial and compositional challenges\\
    \midrule
    HQ-Edit & 500 & High-resolution images with rich details and comprehensive editing instructions\\
    \bottomrule
  \end{tabular}%
  }
\end{table}

We evaluate on three benchmarks. \textbf{Eval-400} (in-house) contains 400 curated image pairs spanning object manipulation, spatial transformation, and attribute modification. \textbf{ByteMorph-Bench}~\cite{chang2025bytemorph} covers non-rigid motion transformations across five categories (Camera Zoom, Camera Motion, Object Motion, Human Motion, and Interaction), making it a challenging testbed for spatial reasoning. \textbf{HQ-Edit}~\cite{hui2024hq} provides high-quality editing pairs generated via GPT-4V and DALL-E 3, We randomly sample 500 instances for evaluation, focusing on fine-grained description capability and instruction completeness.
\subsubsection{Compared Models and Training details}
We compare our method against both open-source and closed-source multimodal models. \textbf{Open-source models}: Qwen3-VL-32B, Qwen3-VL-235B-A22B, Qwen3.5-397B-A17B, GLM-4.5V and Kimi-K2.5. \textbf{Closed-source models}: Gemini-3-Pro and GPT-4.1. For fair comparison, all models are prompted with the same input format, consisting of the source image, target image, and a unified instruction generation prompt.

Full training hyperparameters are reported in Appendix~\ref{app:training_hyperparameters}; all experiments are trained on 32 H800 GPUs.
\subsection{Evaluation Protocol}
To comprehensively evaluate the quality of generated editing instructions, we adopt both human evaluation and automatic evaluation, capturing complementary aspects of instruction usability and quantitative performance.
\subsubsection{Human Evaluation}
We conduct human evaluation on the in-house test set consisting of 400 image pairs. Annotators are asked to assess the generated instructions using a three-level defect classification scheme (P0/P1/P2), organized in descending order of severity.
\begin{itemize}
\item {\texttt{P0 (Critical Errors)}}: Errors that render the instruction unusable, including incorrect description of the main subject, spatial or viewpoint inconsistencies (e.g., left/right confusion), and omission of key editing actions.
\item {\texttt{P1 (Major Errors)}}: Substantial issues that affect editing quality but do not fully invalidate the instruction, such as incorrect description of secondary elements, missing global visual attributes (e.g., lighting, color tone), or leakage of target-image information.
\item {\texttt{P2 (Minor Issues)}}: Minor imperfections that do not affect executability, such as redundant or slightly imprecise phrasing.
\end{itemize}
Annotators are instructed to follow a hierarchical evaluation protocol: if a P0 error is identified, lower-level categories (P1 and P2) are not considered. This design ensures that evaluation focuses on the most critical failure modes affecting usability. This evaluation primarily reflects the practical usability of generated instructions in real-world editing scenarios.
\subsubsection{Automatic Evaluation}
To enable scalable and fine-grained comparison across models, we introduce an automatic evaluation protocol using Gemini-2.5-Pro as a judge model, combined with human-annotated ground truth (GT).

\textbf{High-Quality GT as Absolute Anchor. }Rather than relying solely on the LLM's linguistic intuition for scoring, we establish a rigorous reference standard. The GT annotations are constructed by aggregating visual understanding outputs from multiple state-of-the-art models, followed by manual verification and correction. Each sample is associated with structured GT annotations, including: primary changes, secondary changes, and overall transformation description. The generated instruction is evaluated along three fully orthogonal dimensions:
\begin{itemize}
\item \textbf{Accuracy}: Whether the instruction is factually consistent 
with the visual transformation, including correctness of object attributes 
and absence of hallucinations. Instructions covering less than 30\% of 
primary changes are penalized.
\item \textbf{Completeness}: Coverage of primary changes, defined as the 
ratio of correctly mentioned primary changes to total primary changes in GT. 
This metric focuses purely on omission, independent of correctness.
\item \textbf{Clarity}: Whether the instruction is clear and executable 
under a ``blind execution'' criterion—i.e., actionable without access to 
the target image. Higher scores reward fine-grained attributes (e.g., color, 
texture, spatial relations) and penalize vague expressions.
\end{itemize}

The overall score is computed as a weighted sum:
\begin{equation}
\resizebox{0.92\columnwidth}{!}{$
\text{Score} = 0.4 \times \text{Accuracy} + 0.4 \times \text{Completeness} + 0.2 \times \text{Clarity}
$}
\end{equation}

All models are evaluated under the same protocol with a fixed evaluation prompt (provided in the appendix), ensuring consistency and fairness.

\textbf{Complementarity of Evaluations}. The two evaluation protocols serve complementary purposes: Human evaluation focuses on instruction usability, identifying critical errors that may invalidate editing, Automatic evaluation enables scalable and fine-grained comparison, supporting quantitative benchmarking across models. Together, they provide a comprehensive assessment of both the practical effectiveness and quantitative performance of instruction generation.
\subsection{Results And Analysis}
\begin{table}[t]
\centering
\caption{
Human evaluation results on 400 image pairs. Each model generates one instruction per pair. P0: critical errors (instruction unusable); P1: moderate errors; P2: minor issues. Higher Correct and lower  P0 error rates indicate better performance.
}
\label{tab:human}
\resizebox{\columnwidth}{!}{%
\begin{tabular}{lcccc}
\toprule
\textbf{Model} & \textbf{Correct} $\uparrow$ & \textbf{P0} $\downarrow$ & \textbf{P1} & \textbf{P2}  \\
\midrule
\multicolumn{5}{l}{\textit{Closed-source models}} \\
Gemini-3-Pro          & \textbf{66.00\%}  & 21.00\%  & 12.00\%  & 1.00\%  \\
GPT-4.1               & 48.75\%  & 42.25\%  & 8.25\%   & 0.75\%  \\
\midrule
\multicolumn{5}{l}{\textit{Open-source baseline models}} \\
GLM-4.5V (106B-A12B)      & 38.60\%  & 51.20\%  & 9.20\%   & 1.00\%  \\
Qwen3-VL-235B-A22B (base) & 41.75\%  & 47.75\%  & 8.50\%   & 2.00\%  \\
Qwen3-VL-32B (base)       & 37.00\%  & 51.50\%  & 11.00\%  & 0.50\%  \\
\midrule
\multicolumn{5}{l}{\textit{Our fine-tuned models}} \\
Qwen3-VL-32B (SFT+DPO)$^\dagger$         & 57.00\% & 29.00\% & 13.25\% & 0.75\% \\
Qwen3-VL-32B (SFT+HAE-DPO)$^\dagger$     & 61.25\% & 23.75\% & 14.25\% & 0.75\% \\
Qwen3-VL-235B-A22B (SFT+DPO)$^\dagger$   & 66.00\% & 23.00\% & 10.40\% & 0.60\% \\
Qwen3-VL-235B-A22B (SFT+HAE-DPO)$^\dagger$ & \textbf{70.25\%} & \textbf{17.50\%} & 11.50\% & 0.75\% \\
\bottomrule
\end{tabular}%
}
\end{table}
\begin{table*}[t]
\centering
\caption{
Objective evaluation results on three benchmarks.
Scores are reported as weighted composites
($S = 0.4 \times \text{Acc} + 0.4 \times \text{Comp} + 0.2 \times \text{Clar}$, scale 0--5). Best and second-best scores are highlighted in bold and underline, respectively. $\dagger$ marks our fine-tuned models.
}
\label{tab:objective}
\resizebox{\textwidth}{!}{%
\begin{tabular}{l cccc cccc cccc}
\toprule
& \multicolumn{4}{c}{\textbf{Eval-400 (In-house)}}
& \multicolumn{4}{c}{\textbf{HQ-Edit}}
& \multicolumn{4}{c}{\textbf{ByteMorph-Bench}} \\
\cmidrule(lr){2-5} \cmidrule(lr){6-9} \cmidrule(lr){10-13}
\textbf{Model}
& $S$ & Acc & Comp & Clar
& $S$ & Acc & Comp & Clar
& $S$ & Acc & Comp & Clar \\
\midrule
\multicolumn{13}{l}{\textit{Closed-source models}} \\
Gemini-3-Pro
& \underline{4.706} & 4.70 & 4.85 & 4.43
& \underline{4.658} & 4.54 & 4.89 & 4.43
& 4.522 & 4.55 & 4.70 & 4.11 \\
GPT-4.1
& 4.220 & 4.03 & 4.60 & 3.84
& 4.507 & 4.57 & 4.86 & 3.68
& 3.412 & 3.27 & 3.68 & 3.14 \\
\midrule
\multicolumn{13}{l}{\textit{Open-source baseline models}} \\
Qwen3.5-397B-A17B
& 4.380 & 4.30 & 4.62 & 4.06
& 4.383 & 4.07 & 4.85 & 4.07
& 3.867 & 3.71 & 4.16 & 3.61 \\
Kimi-K2.5
& 4.111 & 3.69 & 4.72 & 3.74
& 4.310 & 3.89 & 4.94 & 3.90
& 3.679 & 2.94 & 4.57 & 3.38 \\
GLM-4.5V (106B-A12B)
& 3.970 & 4.36 & 4.33 & 3.52
& 3.384 & 3.64 & 3.23 & 3.19
& 3.448 & 3.25 & 3.78 & 3.19 \\
Qwen3-VL-32B
& 3.480 & 3.14 & 3.60 & 2.92
& 4.007 & 3.69 & 4.61 & 3.62
& 3.332 & 3.06 & 3.67 & 3.20 \\
Qwen3-VL-235B-A22B
& 3.880 & 3.65 & 3.78 & 3.48
& 4.397 & 4.30 & 4.88 & 3.63
& 3.462 & 3.19 & 3.90 & 3.13 \\
\midrule
\multicolumn{13}{l}{\textit{Our fine-tuned models}} \\
Qwen3-VL-32B (SFT)$^\dagger$
& 4.349 & 4.30 & 4.58 & 3.98
& 4.387 & 4.12 & 4.76 & 4.17
& 3.914 & 3.78 & 4.08 & 3.84 \\
Qwen3-VL-32B (SFT+DPO)$^\dagger$
& 4.386 & 4.31 & 4.63 & 4.05
& 4.458 & 4.22 & 4.80 & 4.26
& 3.931 & 3.73 & 4.17 & 3.86 \\
Qwen3-VL-32B (SFT+HAE-DPO)$^\dagger$
& 4.456 & 4.40 & 4.69 & 4.10
& 4.520 & 4.30 & 4.86 & 4.28
& 4.068 & 3.85 & 4.31 & 4.02 \\
Qwen3-VL-235B-A22B (SFT)$^\dagger$
& 4.521 & 4.44 & 4.79 & 4.15
& 4.552 & 4.40 & 4.88 & 4.12
& 4.208 & 4.07 & 4.51 & 3.89 \\
Qwen3-VL-235B-A22B (SFT+DPO)$^\dagger$
& 4.662 & 4.63 & 4.85 & 4.34
& 4.598 & 4.46 & 4.89 & 4.29
& \underline{4.564} & 4.68 & 4.72 & 4.02 \\
Qwen3-VL-235B-A22B (SFT+HAE-DPO)$^\dagger$
& \textbf{4.720} & 4.72 & 4.89 & 4.38
& \textbf{4.672} & 4.61 & 4.87 & 4.40
& \textbf{4.651} & 4.75 & 4.79 & 4.17 \\
\bottomrule
\end{tabular}%
}
\end{table*}

\begin{table}[t]
\centering
\caption{
Ablation study on HAE-DPO margin components using Qwen3-VL-235B-A22B with the SFT stage fixed.
The term $s_i c(t_i)$ denotes the severity-type product, where $s_i$ is the error severity and $c(t_i)$ is the failure-type weight.
The term $h_i$ denotes model-perceived hardness.
Rows ``Only $s_i c(t_i)$'' and ``Only $h_i$'' add the corresponding margin component alone, while HAE-DPO uses both.
$S = 0.4{\times}$Acc $+ 0.4{\times}$Comp $+ 0.2{\times}$Clar.}
\label{tab:ablation}
\scriptsize
\setlength{\tabcolsep}{2.2pt}
\renewcommand{\arraystretch}{0.92}
\resizebox{\columnwidth}{!}{%
\begin{tabular}{lcccccccccccc}
\toprule
& \multicolumn{4}{c}{\textbf{Eval-400}}
& \multicolumn{4}{c}{\textbf{HQ-Edit}}
& \multicolumn{4}{c}{\textbf{ByteMorph}} \\
\cmidrule(lr){2-5}\cmidrule(lr){6-9}\cmidrule(lr){10-13}
\textbf{Components}
& $S$ & Acc & Comp & Clar
& $S$ & Acc & Comp & Clar
& $S$ & Acc & Comp & Clar \\
\midrule
None (Standard DPO)
& 4.662 & 4.63 & 4.85 & 4.34
& 4.598 & 4.46 & 4.89 & 4.29
& 4.564 & 4.68 & 4.72 & 4.02 \\
\midrule
Only $s_ic(t_i)$ (severity-type)
& 4.698 & 4.70 & 4.86 & 4.37
& 4.648 & 4.55 & 4.90 & 4.34
& 4.612 & 4.73 & 4.75 & 4.10 \\
\midrule
Only $h_i$ (hardness)
& 4.710 & 4.71 & 4.88 & 4.37
& 4.664 & 4.58 & 4.89 & 4.38
& 4.630 & 4.74 & 4.77 & 4.13 \\
\midrule
\textbf{HAE-DPO} 
& \textbf{4.720} & 4.72 & 4.89 & 4.38
& \textbf{4.672} & 4.61 & 4.87 & 4.40
& \textbf{4.651} & 4.75 & 4.79 & 4.17 \\
\bottomrule
\end{tabular}%
}
\end{table}

\paragraph{Human evaluation.}
Table~\ref{tab:human} presents human evaluation results. Our fine-tuned
Qwen3-VL-235B-A22B achieves a correct rate of 70.25\% and
reduces P0 errors from 47.75\% to 17.50\%, surpassing Gemini-3-Pro's
performance (66\% correct, 21\% P0). The 32B model also shows
significant gains (37\%→61.25\% correct, 51.5\%→23.75\% P0).
These improvements demonstrate that our two-stage training
effectively suppresses critical failure modes such as spatial
relation confusion and subject attribute errors. The slight
increase in P1 errors stems from more detailed descriptions
that occasionally introduce minor inaccuracies, while
significantly reducing critical P0 errors.
\paragraph{Objective evaluation.}
As shown in Table~\ref{tab:objective}, Qwen3-VL-235B-A22B with SFT+HAE-DPO achieves the best overall scores across the three benchmarks, with 4.720 on Eval-400, 4.672 on HQ-Edit, and 4.651 on ByteMorph-Bench. Compared with standard DPO, HAE-DPO further improves the 235B model from 4.662 to 4.720 on Eval-400 and from 4.564 to 4.651 on ByteMorph-Bench. It also surpasses Gemini-3-Pro under our automatic evaluation protocol on three benchmarks.

\paragraph{Ablation study.}
Table~\ref{tab:ablation} ablates the two margin components in HAE-DPO, namely the severity-type product $s_i c(t_i)$ and the model-perceived hardness term $h_i$. 
All variants share the same SFT model, so the comparison isolates the effect of the DPO margin design. 
Compared with standard DPO, adding only $s_i c(t_i)$ consistently improves the overall score $S$ on Eval-400, HQ-Edit, and ByteMorph, indicating that the product of error severity and failure-type weight provides a more fine-grained preference signal. 
Adding only $h_i$ further improves performance, showing that model-perceived hardness is also beneficial for margin calibration. 
The full HAE-DPO model, which combines both components, achieves the best $S$ on all three datasets, with gains of 0.058, 0.074, and 0.087 over standard DPO. 
This demonstrates that the two components are complementary.

Overall, these results demonstrate that our approach not only improves absolute performance, but also systematically enhances structured understanding of visual transformations. Qualitative comparisons are provided in Appendix~\ref{fig:showcase},
demonstrating that our model consistently produces more spatially
accurate and fine-grained instructions than baseline models.

\paragraph{Downstream Editing Quality.}
To assess whether higher-quality instructions translate to better editing
performance, we fine-tune Step1X-Edit~\cite{liu2025step1x} on 1M
image-editing triplets annotated by two captioning sources: Qwen3-VL-235B
(base, before post-training) and our EditCaption model. Fine-tuning is
conducted on 64 H800 GPUs for three days. As shown in
Table~\ref{tab:downstream}, EditCaption-annotated data consistently
outperforms the base VLM baseline across both
ImgEdit-Bench~\cite{ye2026imgedit} and GEdit-Bench~\cite{liu2025step1x}
(EN and CN), confirming that instruction quality directly governs downstream
editing fidelity.

\begin{table}[t]
\centering
\small
\caption{Downstream image editing evaluation on ImgEdit-Bench and GEdit-Bench
(EN and CN). All models are fine-tuned from Step1X-Edit on 1M triplets annotated
by different captioning sources. Scores are overall ratings evaluated by GPT-4.1.}
\label{tab:downstream}
\setlength{\tabcolsep}{1.8pt}
\begin{tabular}{lccc}
\toprule
Model & ImgEdit $\uparrow$ & GEdit-EN $\uparrow$ & GEdit-CN $\uparrow$ \\
\midrule
Step1X-Edit                         & 3.062 & 6.444 & 6.658 \\
+ Base VLM captioner          & 3.358 & 6.583 & 6.774 \\
+ EditCaption \textbf{(ours)} & \textbf{3.882} & \textbf{7.158} & \textbf{7.312} \\
\bottomrule
\end{tabular}
\end{table}

\section{Conclusion And Future Work}
We present \textbf{EditCaption}, a two-stage post-training pipeline for
editing instruction synthesis, comprising 100K human-refined SFT data and
10K human-annotated preference pairs. We further introduce
\textbf{HAE-DPO}, a failure-mode-aware preference optimization method that
modulates the preference margin according to error severity, failure type,
and model-perceived difficulty. Experiments on three benchmarks show that
our fine-tuned Qwen3-VL models significantly outperform open-source and closed-source models, with human evaluation
confirming a 30.25-point reduction in critical errors and a 28.50-point
improvement in correct rate. Downstream fine-tuning of Step1X-Edit on
EditCaption-annotated data further validates that instruction quality
directly governs editing fidelity. Our work demonstrates that targeted, failure-mode-driven alignment provides
a practical and scalable approach for instruction-image data construction.

Despite its effectiveness, our approach has several limitations. First, the model achieves around 70\% accuracy in human
evaluation, which is not yet reliable enough for fully
autonomous large-scale data construction. The current
pipeline depends on external filtering mechanisms such
as EditScore, introducing additional complexity and
highlighting a fundamental constraint on data quality.
Future work could explore integrated quality-aware
generation to reduce reliance on post-hoc filtering. Second, the model may struggle with complex cases such
as multi-step edits, rare transformations, or highly
abstract instructions, due to limitations in both
training data coverage and model capability. Future work
may explore more diverse data construction strategies
and improved modeling of compositional transformations.

\bibliography{custom}

\appendix
\section{Human Evaluation Details}

We provide a detailed summary of the human evaluation protocol in Table~\ref{tab:checklist}, including the definitions of P0, P1, and P2 error categories, along with annotation guidelines and representative examples to ensure consistent assessment.

\begin{table*}[!tbp]
\centering
\caption{Annotation Quality Checklist for Image Editing Instructions}
\label{tab:checklist}
\resizebox{\textwidth}{!}{%
\begin{tabular}{c c p{3cm} p{5cm} p{6cm}}
\toprule
\textbf{ID} & \textbf{Severity} & \textbf{Category} & \textbf{Problem Description} & \textbf{Examples / Notes} \\
\midrule

1
& P0 & (Subject) Text Recognition / Editing Error
& Instruction conflicts with text in image; text unchanged, incorrectly modified, garbled, or misspelled.
& Instruction: ``change sign to `OPEN'\,'', but target still shows ``SHOP''; or text becomes distorted/garbled. \\
\cmidrule(l){2-5}
& P1 & (Background) Text Recognition / Editing Error
& Same as above. & Same as above. \\
\midrule

2
& P0 & (Subject) Instruction or Attribute Error
& Primary task instruction or attribute is incorrect; or affected region exceeds 1/10 of total image area. Covers: color, shape, material, style, quantity, etc.
& ``Change black skirt to blue'' but result is grey; ``Change two cats to three'' but target still has two; ``Enlarge to 15cm'' or ``Rotate 37 degrees'' (unverifiable from image). \\
\cmidrule(l){2-5}
& P1 & (Non-subject) Instruction or Attribute Error
& Non-primary task; affected area below 1/10 of image. Same instruction types as above.
& Same as above. \\
\midrule

3
& P0 & (Subject) Instruction or Content Omission
& Primary task instruction or attribute is missing. Includes: missing add/remove action; missing specific attributes (color, position) when adding/deleting; ambiguous editing target.
& Original has no hat $\to$ target has hat, but ``add hat'' not mentioned; ``Add flower'' without specifying color; ``Delete person'' when multiple people are present; ``Remove old flower + add new flower'' should use \textit{replace}; ``Put object somewhere'' should use \textit{move} not \textit{add}; missing description of border/frame edits. \\
\cmidrule(l){2-5}
& P1 & (Non-subject) Instruction or Content Omission
& Non-primary task; affected area below 1/10 of image. Same instruction types as above.
& Same as above. \\
\midrule

4 & P0 & Viewpoint and Spatial Relation Error
& Applies only within the same scene (not scene replacement). Errors or omissions in: shooting angle, orientation, distance, or occlusion relationships.
& ``Change to holding cup with right hand'' but target shows left hand. \\
\midrule

5 & P1 & Global Visual Attribute Error
& Incorrect or missing changes in lighting, tone, brightness, or saturation.
& Image becomes noticeably brighter but this change is not mentioned. \\
\midrule

6 & P1 & Information Leakage (Target Leak)
& Instruction contains information only obtainable by viewing the target image (e.g., referencing target-specific features).
& Image A has a cup, Image B does not; B$\to$A: ``Put the cup back in place'' — this leaks target information. \\
\midrule

7 & P2 & Redundant or Verbose Description
& Repetitive, contradictory, or meaningless descriptions.
& ``Keep white unchanged, change to white''; ``Remove red flower, add blue flower at same location'' should be ``Replace red flower with blue flower''. \\
\midrule

8 & P2 & Other Undefined Issues
& Other minor or unstructured issues; incorrect description of two identical images.
& Add remarks as needed. \\
\bottomrule
\end{tabular}%
}
\end{table*}

\section{Automatic Evaluation Details}

\label{sec:appendix-prompts}

We evaluate generated image editing instructions along three
dimensions: \textit{Accuracy}, \textit{Completeness}, and
\textit{Clarity}. Each prompt is provided to a multimodal
judge model together with the source image (Image~A), the edited
image (Image~B), and the ground truth change descriptions.
The final score is computed as a weighted sum:
$S = 0.4 \cdot S_{\text{acc}} + 0.4 \cdot S_{\text{com}} + 0.2 \cdot S_{\text{cla}}$.
\subsection{Accuracy Evaluation Prompt}
\label{sec:prompt-accuracy}
The prompt used for evaluating the Accuracy dimension is provided in Listing~\ref{lst:prompt-accuracy}. This prompt guides the evaluator to assess whether the generated instruction is factually consistent with the transformation between the source and target images, with particular emphasis on detecting hallucinations, incorrect attribute descriptions, and spatial inconsistencies.
\begin{lstlisting}[style=promptstyle,
  caption={Accuracy Evaluation Prompt},
  label={lst:prompt-accuracy}]
[SYSTEM] You are evaluating ONLY the [Accuracy] dimension.

Key Reminders:
- Even if an instruction is complete or clear, any factual
  error MUST result in score deduction.
- Even if an instruction is incomplete or vague, if all
  stated content is factually correct, a high score may
  be given.
- Do NOT let Completeness or Clarity influence Accuracy.

Core Principle: Lie Detector Mode + Minimum Coverage.
Accuracy = factual correctness + minimum coverage threshold.
- Omitting attribute degree (GT: "deep red" -> "red") is a
  Clarity issue; it does NOT affect Accuracy.
- At least 30-50% of major changes must be covered to
  receive a high score.

[Ground Truth]
{gt_text}

[Instruction to Evaluate ({model_name})]
{instruction}

Evaluation Steps (output each step):
  Step 1: Count total GT major changes (N).
  Step 2: List all assertions in the instruction.
  Step 3: Hallucination check -- any hallucination -> max 2.
  Step 4: Fact-check each assertion (object / op / attribute).
  Step 5: Coverage rate = M / N (M: major changes hit).
  Step 6: Determine score via rubric (errors + coverage).
  Step 7: Output as single-line JSON (no markdown fences):
    {"dimension":"accuracy","score":<int>,"reasoning":"..."}
\end{lstlisting}

\subsection{Completeness Evaluation Prompt}
\label{sec:prompt-completeness}
The prompt for the Completeness dimension is shown in Listing~\ref{lst:prompt-completeness}. It focuses on measuring the coverage of primary changes by comparing the generated instruction with the annotated ground truth, encouraging the evaluator to identify missing key transformations regardless of correctness.
\begin{lstlisting}[style=promptstyle,
  caption={Completeness Evaluation Prompt},
  label={lst:prompt-completeness}]
[SYSTEM] You are evaluating ONLY the [Completeness] dimension.

Goal: Compute recall rate of the instruction vs. ground truth.

Core Principles:
  1. Count only: check whether each GT change is "triggered"
     in the instruction -- correctness does not matter here.
  2. No double-penalization: score is set by coverage rate
     alone; do not further deduct after computing coverage.

Hit / Miss Definition:
  HIT:  Instruction mentions the GT object + action type.
        (Wrong attributes -> Hit; vague expression -> Hit)
  MISS: GT change not mentioned at all, OR action direction
        is reversed (GT: add / instruction: remove).

Coverage Rate: R = K / N x 100%
  (K = hits, N = total GT major changes)

Score Lookup Table (strict; no arbitrary adjustment):
  R = 100%        -> 5.0   75% <= R < 100% -> 4.5
  60% <= R < 75%  -> 4.0   50% <= R < 60%  -> 3.0
  20% <= R < 50%  -> 2.0   R = 0%          -> 1.0

Bonus (only allowed upward adjustment):
  If R < 100% but GT minor changes are additionally covered:
  +0.5. Downward adjustments are strictly PROHIBITED.

[Ground Truth]
{gt_text}

[Instruction to Evaluate ({model_name})]
{instruction}

Evaluation Steps (output each step):
  Step 1: List all GT major changes; label each Hit/Miss.
  Step 2: Compute R = K / N.
  Step 3: Look up base score in table.
  Step 4: Check minor change bonus (+0.5 if applicable).
  Step 5: Output as single-line JSON (no markdown fences):
    {"dimension":"completeness","score":<float>,
     "reasoning":"N changes, K hits (R=X%)..."}
\end{lstlisting}
\subsection{Clarity Evaluation Prompt}
\label{sec:prompt-clarity}
The prompt used to evaluate Clarity is presented in Listing~\ref{lst:prompt-clarity}. It follows a “blind execution” criterion, assessing whether the instruction is sufficiently precise, unambiguous, and detailed for a user to reproduce the intended edit without access to the target image.
\begin{lstlisting}[style=promptstyle,
  caption={Clarity Evaluation Prompt},
  label={lst:prompt-clarity}]
[SYSTEM] You are evaluating ONLY the [Clarity] dimension.

Key Reminders:
- Even if complete and accurate, vague expression MUST
  result in deduction.
- Even if incorrect or incomplete, clear expression can
  still earn a high score.
- Do NOT let Completeness or Accuracy influence Clarity.

Core Principle: Executability Test + Clarity-First.
Test: could a "blind" robot execute this instruction
precisely from text alone?

Three Criteria (by importance):
  1. Unambiguity      (50%): no forbidden terms; operation
                             and target state clearly defined.
  2. Attr. Completeness(40%): color/material/shape/size
                             explicitly described.
  3. Info. Density    (10%): concise > verbose; >150 words
                             with <5 info points -> deduct.

Forbidden Terms (deduct if any present):
  Vague verbs:    adjust / process / optimize / modify
                  (without explicit target state)
  Vague degrees:  appropriate / slightly / a bit / somewhat
  Vague refs:     that / this / it / refer to original image

Forbidden Term Score Ceiling:
  1 term  -> max 3.0 | 2 terms -> max 2.5 | 3+ -> max 2.0

[Ground Truth (attribute check only)]
{gt_changes}

[Instruction to Evaluate ({model_name})]
{instruction}

Evaluation Steps (output each step):
  Step 1: Detect forbidden terms (list each found).
  Step 2: Check attribute completeness per object.
  Step 3: Assess detail level (location/angle/secondary attr).
  Step 4: Set score ceiling by forbidden term count.
  Step 5: Adjust by attribute completeness and detail.
  Step 6: Output as single-line JSON (no markdown fences):
    {"dimension":"clarity","score":<float>,"reasoning":"..."}
\end{lstlisting}
\subsection{Qualitative Examples}
\label{subsec:qualitative}
\begin{figure*}[!tbp]
  \centering
  \includegraphics[width=\linewidth]{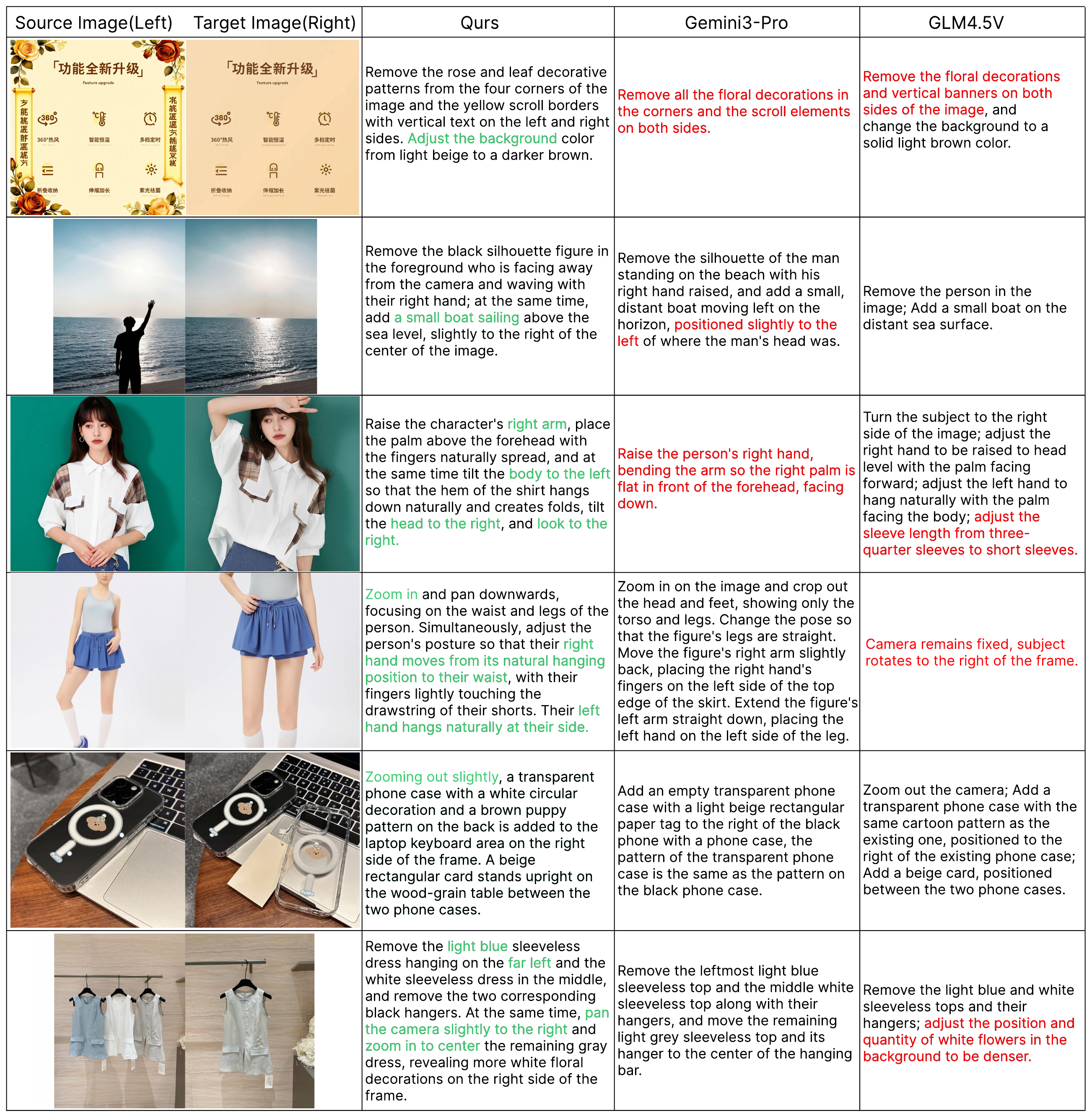}
  \caption{
    \textbf{Qualitative examples of generated editing instructions.}
    Given the same source-target image pairs, our model produces instructions
    that are more spatially precise and complete compared to Gemini-3-Pro and GLM4.5V.
    \textbf{Green} denotes key attributes where our model
  outperforms baselines; \textbf{Red} denotes errors or omissions.
  }
  \label{fig:showcase}
\end{figure*}

Figure~\ref{fig:showcase} presents side-by-side comparisons of
editing instructions generated by our model, Gemini-3-Pro, and GLM4.5V
on diverse source-target image pairs. Our model consistently captures
directional relations, viewpoint changes, and fine-grained attributes
that baseline models frequently omit or misidentify.

\subsection{Scoring Examples}
To provide a concrete and holistic understanding of the evaluation process, we present a unified case study based on a single source-target image pair and its annotated ground truth (GT).
We then illustrate how the generated instruction is evaluated across different dimensions, including Accuracy, Completeness, and Clarity, with detailed reasoning for each score.
\subsubsection{Ground Truth Example}
We first present the ground truth annotation for the selected example in Table~\ref{tab:annotation-example}, including primary changes, secondary changes, and an overall description of the transformation.

\begin{table}[!tbp]
\centering
\caption{An annotation example of image editing instructions.}
\label{tab:annotation-example}
\scalebox{0.78}{%
\begin{tabular}{|c|p{0.55\columnwidth}|}
\hline
\textbf{Images} & \textbf{Annotation} \\
\hline
\begin{minipage}{3.2cm}
  \centering
  \includegraphics[width=\linewidth]{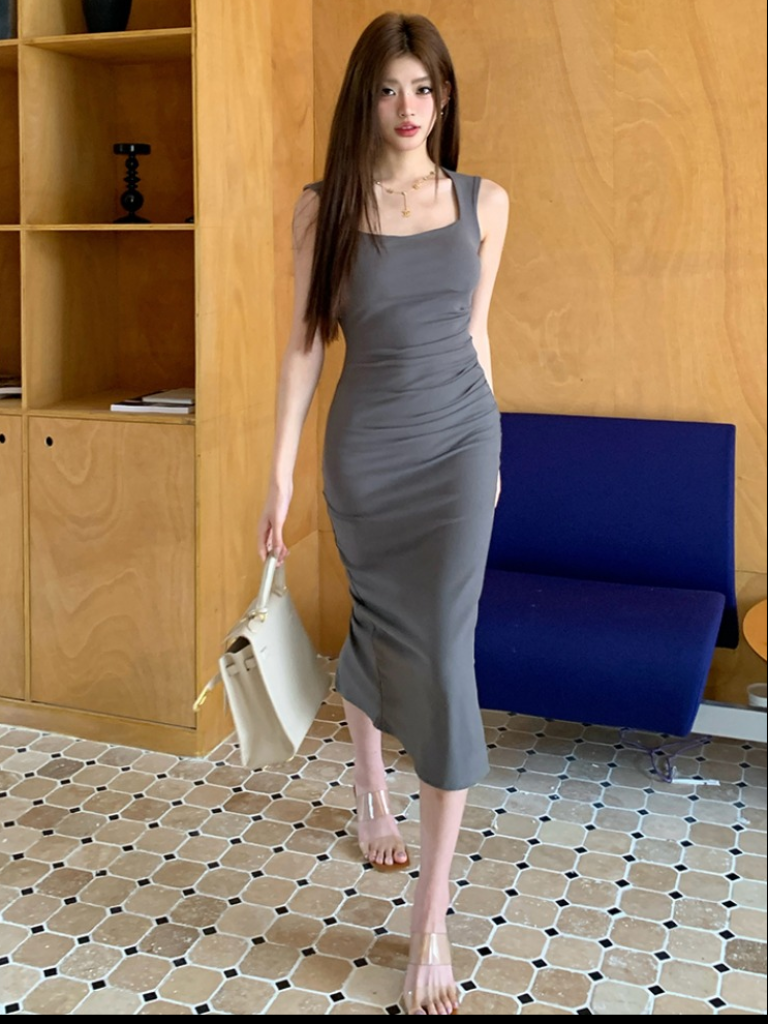}
  \vspace{0.3em}
  \includegraphics[width=\linewidth]{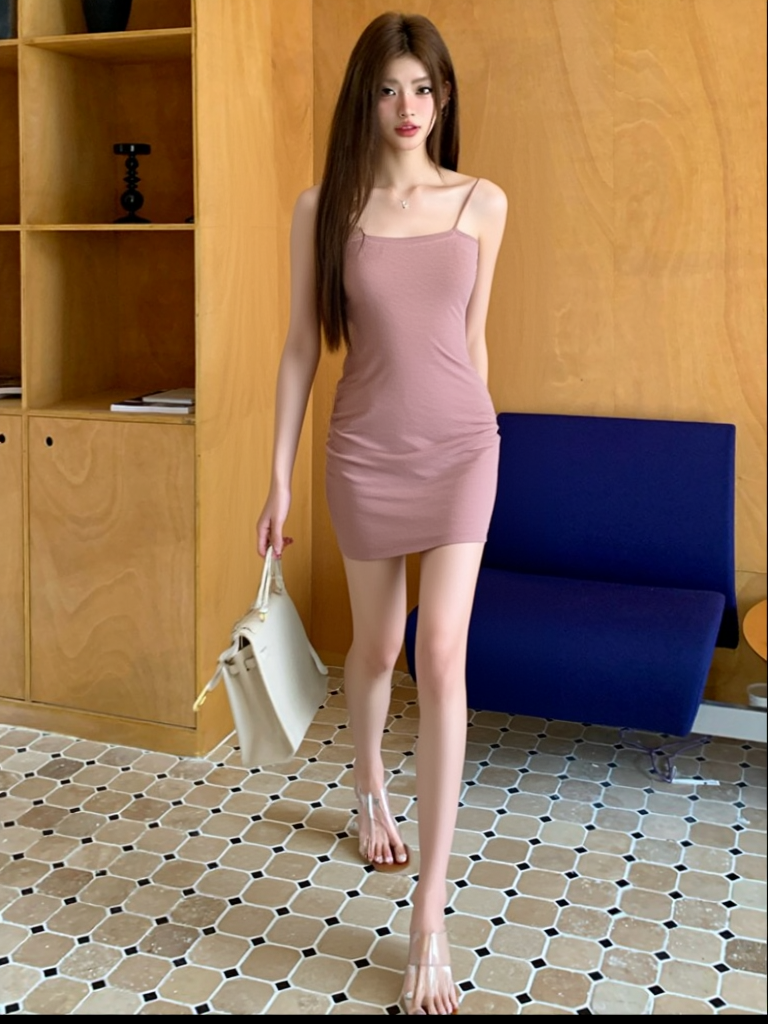}
\end{minipage}
&
\begin{minipage}[c]{0.53\columnwidth}
  \vspace{0.3em}
  \normalsize
  \textbf{[Primary Changes]}
  \begin{enumerate}[leftmargin=*, nosep]
    \item The dress color changes from grey to pink.
    \item The shoulder strap style changes from wide straps
          to thin spaghetti straps.
    \item The dress hem length is shortened from midi-length
          (approximately mid-calf) to a mini skirt
          (approximately mid-thigh).
  \end{enumerate}
  \textbf{[Secondary Changes]}
  \begin{enumerate}[leftmargin=*, nosep]
    \item The gold necklace around the neck is replaced
          with a thin silver necklace.
  \end{enumerate}
  \textbf{[Overall Description]}

  \noindent The woman in the image, originally wearing a grey
  wide-strap midi dress, has been edited to wear a pink
  spaghetti-strap mini dress. Her gold necklace has been
  replaced with a thin silver one. Her pose and white handbag
  remain unchanged.
  \vspace{0.3em}
\end{minipage}
\\
\hline
\end{tabular}
}%
\end{table}

\subsubsection{Accuracy Evaluation}
We next demonstrate how the Accuracy score is assigned under the “fact-checking” criterion. Table~\ref{tab:scoring_rubric} presents representative instructions with different scores, along with detailed reasoning.

\subsubsection{Completeness Evaluation}
The Completeness score is computed based on the coverage of primary changes in the ground truth. Table~\ref{tab:scoring_coverage} presents representative instructions with different scores, along with detailed reasoning.

\subsubsection{Clarity Evaluation}
We further illustrate the evaluation of Clarity, which measures whether an instruction is sufficiently detailed and unambiguous under a blind execution setting. Table~\ref{tab:scoring_specificity} presents representative instructions with different scores, along with detailed reasoning.

\begin{table*}[!tbp]
\centering
\scriptsize
\setlength{\tabcolsep}{4pt}
\renewcommand{\arraystretch}{0.95}
\caption{Representative scoring examples for the Accuracy dimension, with corresponding instructions and reasoning.}
\label{tab:scoring_rubric}
\begin{tabular}{>{\centering\arraybackslash}p{1.2cm} p{4.8cm} p{4.2cm} p{5.2cm}}
\toprule
\textbf{Score} & \textbf{Criteria} & \textbf{Case} & \textbf{Reasoning} \\
\midrule

5 &
\textbf{Perfect accuracy}
\begin{itemize}[leftmargin=*,noitemsep,topsep=2pt]
  \item All facts correct
  \item Coverage $\geq$50\%
\end{itemize}
&
\textit{``Change dress to pink spaghetti-strap; necklace to silver.''}
&
\textbf{All facts correct: }pink \checkmark, spaghetti-strap \checkmark, silver necklace \checkmark. Coverage 2/3 (67\%) $\geq$ 50\%.

\textbf{Note:} dress length not mentioned, but accuracy is perfect and coverage is sufficient.
\\

\midrule

4 &
\textbf{Core correct, minor flaw}
\begin{itemize}[leftmargin=*,noitemsep,topsep=2pt]
  \item Major facts correct
  \item Coverage $\geq$30\%
  \item Error on unchanged elements
\end{itemize}
&
\textit{``Change dress to pink; bag to red.''}
&
Dress pink \checkmark.

Bag unchanged but turned red $\times$.

Non-core error: $-$1.

\\

\midrule

3 &
\textbf{Single error or conservative penalty}
\begin{itemize}[leftmargin=*,noitemsep,topsep=2pt]
  \item 1 major factual error
  \item Or coverage $<$30\%
\end{itemize}
&
\textbf{A:} \textit{``Dress to blue.''}

\smallskip
\textbf{B:} \textit{``Necklace to silver.''}
&
\textbf{A:} GT is pink; blue $\times$.

\smallskip
\textbf{B:} Dress coverage $=0$ $\times$.

Prevents skipping primary edits.
\\

\midrule

2 &
\textbf{Multiple errors or hallucination}
\begin{itemize}[leftmargin=*,noitemsep,topsep=2pt]
  \item $\geq$2 major errors
  \item Or fabricated operations
\end{itemize}
&
\textbf{A:} \textit{``Long blue dress.''}

\smallskip
\textbf{B:} \textit{``Pink dress; add hat.''}
&
\textbf{A:} Color $\times$ + length $\times$.

\smallskip
\textbf{B:} ``Add hat'' not in GT $\times$.

Severely misleads.
\\

\midrule

1 &
\textbf{Completely wrong}
\begin{itemize}[leftmargin=*,noitemsep,topsep=2pt]
  \item Wrong object
  \item Unrelated to image
\end{itemize}
&
\textit{``Change car to red.''}
&
No car in image $\times$.
\\

\bottomrule
\end{tabular}

\vspace{0.2em}
\caption{Representative scoring examples for the Completeness dimension, with corresponding instructions and reasoning.}
\label{tab:scoring_coverage}
\begin{tabular}{>{\centering\arraybackslash}p{1.2cm} p{4.8cm} p{4.2cm} p{5.2cm}}
\toprule
\textbf{Score} & \textbf{Criteria} & \textbf{Case} & \textbf{Reasoning} \\
\midrule

5 &
\textbf{High coverage}
\begin{itemize}[leftmargin=*,noitemsep,topsep=2pt]
  \item Coverage $\geq$90\%
  \item Nearly all major changes covered
\end{itemize}
&
``Change to a pink spaghetti-strap mini dress.''
&
\textbf{Full coverage:}

1.~color (pink) \checkmark; 2.~style (spaghetti-strap) \checkmark;

3.~length (mini) \checkmark. Coverage: 3/3 = 100\%.
\\

\midrule

4 &
\textbf{Medium-high coverage}
\begin{itemize}[leftmargin=*,noitemsep,topsep=2pt]
  \item Coverage $\geq$70\% 
  \item 1 major change missed or partially described
\end{itemize}
&
``Change the dress to pink and shorten it.''
&
1.~color (pink) \checkmark; 2.~style (not mentioned) $\times$;

3.~length (shorten) \checkmark. Coverage: 2/3 = 66\%.
\\

\midrule

3 &
\textbf{Passing coverage}
\begin{itemize}[leftmargin=*,noitemsep,topsep=2pt]
  \item Coverage $\geq$50\%
  \item At least 1 major change mentioned
\end{itemize}
&
``Change the dress to pink.''
&
1.~color (pink) \checkmark;

2.~style (not mentioned) $\times$;

3.~length (not mentioned) $\times$.

Coverage: 1/3 = 33\%.
\\

\midrule

2 &
\textbf{Partial or vague coverage}
\begin{itemize}[leftmargin=*,noitemsep,topsep=2pt]
  \item Coverage $>$30\%
  \item Object mentioned but no specific attribute change stated
\end{itemize}
&
``Change the dress style.''
&
1.~color (not mentioned) $\times$;

2.~style (mentioned but unspecified, counts 0.5) \checkmark;

3.~length (not mentioned) $\times$.

Coverage: 0.5/3 = 16.7\%.
\\

\midrule

1 &
\textbf{Low or no coverage}
\begin{itemize}[leftmargin=*,noitemsep,topsep=2pt]
  \item Coverage $\approx$10\%
  \item Only minor changes mentioned
\end{itemize}
&
``Replace the necklace with a silver one.''
&
\textbf{Avoids primary edits:}

all 3 major changes (dress-related) missed; only a minor change mentioned. Coverage: 0/3 = 0\%.
\\

\bottomrule
\end{tabular}

\vspace{0.2em}
\caption{Representative scoring examples for the Clarity dimension, with corresponding instructions and reasoning.}
\label{tab:scoring_specificity}
\begin{tabular}{>{\centering\arraybackslash}p{1.2cm} p{4.8cm} p{4.2cm} p{5.2cm}}
\toprule
\textbf{Score} & \textbf{Criteria} & \textbf{Case} & \textbf{Reasoning} \\
\midrule

5 &
\textbf{Extremely specific}
\begin{itemize}[leftmargin=*,noitemsep,topsep=2pt]
  \item No ambiguity
  \item Secondary attributes fully described (shade, material, thickness, exact length)
  \item Necessary position or state included
\end{itemize}
&
``Replace the gray wide-strap dress with a pink spaghetti-strap mini dress; change the necklace to a silver thin chain.''
&
\textbf{All attributes present:} color (pink) \checkmark, style (spaghetti-strap) \checkmark, length (mini) \checkmark, necklace (silver thin chain) \checkmark.

Every noun has a specific modifier --- execution result is unambiguous.
\\

\midrule

4 &
\textbf{Detailed and complete}
\begin{itemize}[leftmargin=*,noitemsep,topsep=2pt]
  \item No ambiguity
  \item Core attributes present (color, type)
  \item 1--2 secondary attributes missing
\end{itemize}
&
``Change the dress to a pink strappy short dress; replace the necklace with a silver one.''
&
\textbf{Core attributes clear:} pink \checkmark, strappy \checkmark, short dress \checkmark.

\textbf{Minor gaps:} ``thin'' strap and ``mini'' length not specified; necklace chain type unspecified. Executable, but slightly less precise than score 5.
\\

\midrule

3 &
\textbf{Basically clear but vague}
\begin{itemize}[leftmargin=*,noitemsep,topsep=2pt]
  \item Object and operation identifiable
  \item Key target attributes severely missing
\end{itemize}
&
``Change the dress to pink and swap the necklace.''
&
\textbf{Attributes missing:} only color (pink) \checkmark; style (strappy) $\times$ and length (mini) $\times$ omitted. Necklace change stated but target style and color unspecified.

Execution result is uncertain.
\\

\midrule

2 &
\textbf{Ambiguous or mildly forbidden}
\begin{itemize}[leftmargin=*,noitemsep,topsep=2pt]
  \item 1 forbidden word (e.g., ``slightly'', ``that'')
  \item Or unclear reference
\end{itemize}
&
``Change the color of that piece of clothing to pink.''
&
\textbf{Unclear reference:} ``that piece'' is unspecified; ``clothing'' is broader than ``dress''.

\textbf{Redundant phrasing:} low-information words included. Ambiguous to execute.
\\

\midrule

1 &
\textbf{Not executable}
\begin{itemize}[leftmargin=*,noitemsep,topsep=2pt]
  \item Multiple forbidden words (optimize, process, appropriate)
  \item Or purely subjective description
\end{itemize}
&
``Optimize the image to make it look better.''
&
\textbf{Forbidden words:} ``optimize'', ``look better'' are blacklisted terms.

Purely subjective --- cannot be translated into a concrete operation.
\\

\bottomrule
\end{tabular}
\end{table*}
\section{Training Hyperparameters}
\label{app:training_hyperparameters}

Table~\ref{tab:training_hparams} summarizes the training hyperparameters used in the supervised fine-tuning (SFT) and HAE-DPO stages.
For both stages, we start from the corresponding Qwen3-VL-Instruct checkpoint and train on the constructed image-pair instruction data.
In the HAE-DPO stage, the reference model is fixed to the SFT checkpoint, and the same preference data and hyperparameter settings are used for both model scales unless otherwise specified.
All batch sizes denote the global effective batch size after gradient accumulation.
We use mixed-precision training with bfloat16 and AdamW optimization.

\begin{table}[t]
\centering
\small
\caption{Training hyperparameters for the SFT and HAE-DPO stages. Batch size denotes the global effective batch size after gradient accumulation.}
\label{tab:training_hparams}
\begin{tabular}{lcc}
\toprule
\textbf{Hyperparameter} & \textbf{SFT} & \textbf{HAE-DPO} \\
\midrule
Learning rate              & $1\times10^{-5}$ & $5\times10^{-6}$ \\
Global batch size           & 512               & 128               \\
Training epochs             & 6                & 4                \\
Max sequence length          & 8192             & 8192             \\
Learning-rate schedule       & Cosine decay      & Cosine decay      \\
Warmup ratio                & 0.03             & 0.03             \\
Gradient clipping            & 1.0              & 1.0              \\
Precision                   & bfloat16         & bfloat16         \\
DPO $\beta$                 & --               & 0.1              \\
Hardness temperature $\tau$  & --               & 1.0              \\
$\gamma_{st}, \gamma_h$ & --             & $1.0, 1.0$  \\
Reference model             & --               & Frozen SFT model \\
\midrule
GPU type                    & \multicolumn{2}{c}{H800} \\
Number of GPUs              & \multicolumn{2}{c}{32} \\
Hardware                    & \multicolumn{2}{c}{32$\times$H800} \\
Training time (32B)          & $\sim$30h         & $\sim$7h          \\
Training time (235B)         & $\sim$72h         & $\sim$12h         \\
\bottomrule
\end{tabular}
\end{table}

\end{document}